\documentclass[runningheads]{llncs}

\usepackage{eccv}

\usepackage{eccvabbrv}

\usepackage{graphicx}
\usepackage{booktabs}

\usepackage[accsupp]{axessibility}  

\usepackage{array}
\usepackage{makecell}
\usepackage{siunitx}
\usepackage{wrapfig}
\usepackage{booktabs}

\usepackage{hyperref}

\usepackage{orcidlink}

\newcommand{\blue}[1]{{#1}}
\newcommand{\camready}[1]{{\color{black} #1}}

\usepackage[absolute]{textpos}

\begin{document}

\definecolor{somegray}{rgb}{0.5, 0.5, 0.5}
\newcommand{\darkgrayed}[1]{\textcolor{somegray}{#1}}
\begin{textblock}{8}(4, 0.7)
\begin{center}
\darkgrayed{This paper has been accepted for publication at the \\
European Conference on Computer Vision (ECCV), 2024}
\end{center}
\end{textblock}

\title{Reinforcement Learning Meets Visual Odometry} 

\titlerunning{Reinforcement Learning Meets Visual Odometry}

\author{
Nico Messikommer\thanks{equal contribution} \orcidlink{0000-0003-1444-1176} \and
Giovanni Cioffi$^{\star}$ \orcidlink{0000-0003-3964-8552} \and \\
Mathias Gehrig \orcidlink{0000-0003-2223-4265} \and
Davide Scaramuzza \orcidlink{0000-0002-3831-6778} 
}

\authorrunning{Messikommer et al.}

\institute{Dept. of Informatics, University of Zurich\\
\email{\{nmessi,cioffi,mgehrig,sdavide\}@ifi.uzh.ch}}

\maketitle

\begin{abstract}
Visual Odometry (VO) is essential to downstream mobile robotics and augmented/virtual reality tasks. 
\blue{
Despite recent advances, existing VO methods still rely on heuristic design choices that require several weeks of hyperparameter tuning by human experts, hindering generalizability and robustness.
}
We address these challenges by reframing VO as a sequential decision-making task and applying Reinforcement Learning (RL) to adapt the VO process dynamically. 
Our approach introduces a neural network, operating as an agent within the VO pipeline, to make decisions such as keyframe and grid-size selection based on real-time conditions. 
Our method minimizes reliance on heuristic choices using a reward function based on pose error, runtime, and other metrics to guide the system. 
Our RL framework treats the VO system and the image sequence as an environment, with the agent receiving observations from keypoints, map statistics, and prior poses. 
\blue{
Experimental results using classical VO methods and public benchmarks demonstrate improvements in accuracy and robustness, validating the generalizability of our RL-enhanced VO approach to different scenarios.
We believe this paradigm shift advances VO technology by eliminating the need for time-intensive parameter tuning of heuristics.
}
\end{abstract}    
\begin{figure}[ht!]
\centering
\includegraphics[width=0.7\textwidth]{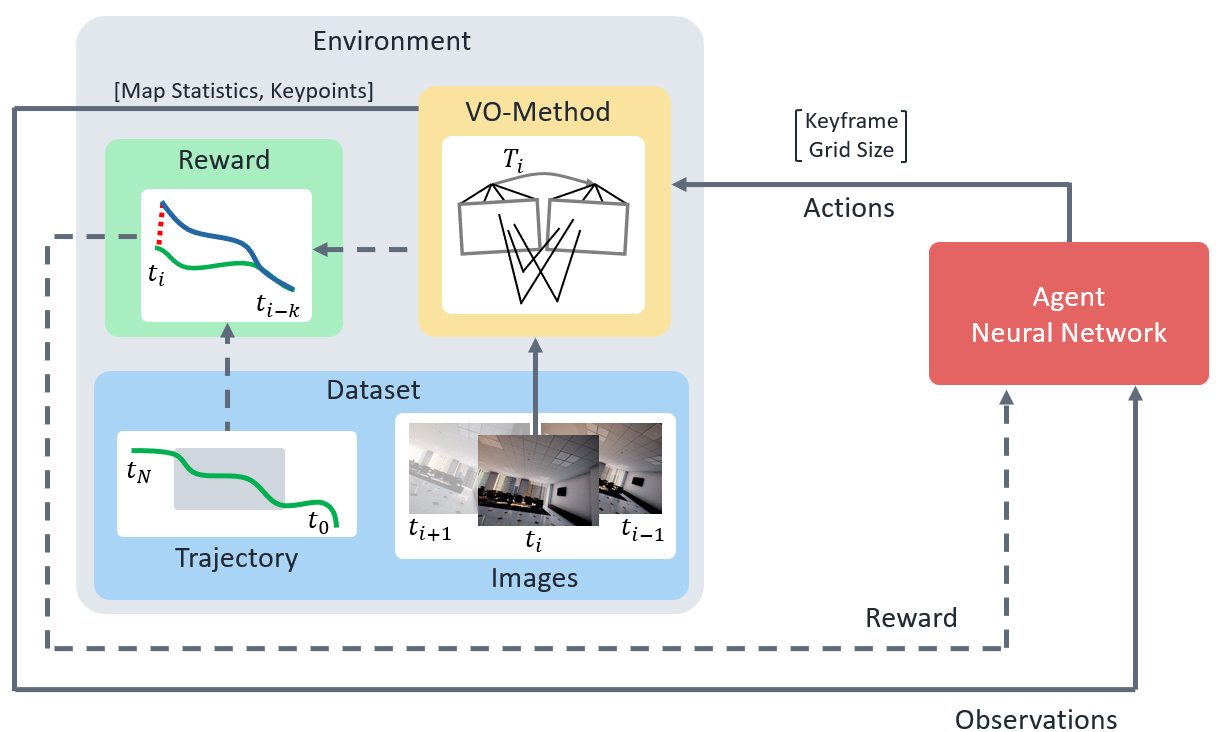}
\caption{
\textbf{Our Framework.}
We propose to employ a learned agent to adaptively guide a VO method using real-time observations for enhanced robustness and accuracy.
By considering the problem as a sequential decision process, we use RL to train the agent primarily based on the position error computed within a sliding window (dashed lines).
}
\label{fig:method_overview}
\end{figure} 

\noindent\textbf{Multimedia Material}
The code is available at \url{https://github.com/uzh-rpg/rl_vo} and video at \url{https://youtu.be/pt6yPTdQd6M}

\section{Introduction}
\label{sec:intro}

The task of estimating the camera pose from a sequence of images, referred to as Visual Odometry (VO), finds broad applications in fields such as robotics and augmented/virtual reality~\cite{scaramuzza2011visual,cadena2016past}. 
\blue{
Despite considerable research and industry efforts to enhance the robustness of VO algorithms, state-of-the-art approaches grapple with generalizability~\cite{cadena2016past,wang20iros}.
This limitation stems from the reliance on hyperparameters dictated by heuristic decision rules, which are highly scenario-dependent, varying with factors such as motion and lighting. 
Consequently, identifying the optimal hyperparameters often requires the intensive effort of domain experts, consuming several weeks.
}
Existing VO methods can be positioned on a spectrum ranging from classical geometry-based approaches to end-to-end methods trained on extensive datasets.
End-to-end approaches~\cite{wang2017deepvo,wang2021tartanvo,ye2023pvo} aim to increase robustness by leveraging a high amount of image sequences with ground truth poses.
However, they tend to struggle to adapt to visual scenarios and motion patterns beyond their training data distribution.
On the other side of the spectrum are classical methods~\cite{Forster17troSVO, Engel17pami, ORBSLAM3_TRO}, which leverage geometry to map image points to and from the 3D world to compute the camera pose.
Hybrid approaches~\cite{teed2021droid, teed2022deep}, positioned between these extremes, replace certain parts of the VO pipeline, usually the ones processing raw image pixels, with learned components while retaining well-defined geometric components to increase robustness.
\blue{Despite these advancements, both hybrid and classical approaches still rely on heuristic design choices, rendering the optimal VO settings scene- and motion-dependent.
Furthermore, these heuristics are defined by a fixed set of hyperparameters found through extensive offline tuning.}
For example, one of the most impactful decisions to be taken by most of the VO systems during operation is the selection of keyframes.
An adequate keyframe selection strategy should add informative keyframes to the current map to maximize the map coverage and, consequently, the accuracy of the registration of the pose of the incoming frame.
Considering the limited map size due to runtime and memory usage constraints, the naive strategy of selecting each frame as a keyframe is oftentimes not the optimal one.
In the current state-of-the-art VO systems~\cite{Forster17troSVO,Engel17pami}, keyframe selection strategies are based on heuristic \textit{if-else} conditions, which rely on a fixed set of hyperparameters.
These hyperparameters are generally tuned offline by expert users, which exploit the prior knowledge of the scene and the expected camera motion.
For example, a strategy prone to often selecting a keyframe is preferred if the camera undergoes fast motion.
Other hyperparameters such as feature extraction thresholds, brightness compensation, and weighting factors are also important, however, their impact on the overall performance depends on the specific VO system.
To reduce the dependence on heuristic-based design, we approach the Visual Odometry problem from a sequential decision-making perspective, where current actions influence future states.
Given the temporal dependencies inherent in this task, coupled with the absence of ground truth for optimal actions, we propose to train a neural network within a Reinforcement Learning (RL) framework to adapt a VO method at each timestep of the sequence.
Operating as a dynamic agent, this neural network guides decisions within the VO pipeline, such as keyframe selection, or, in theory, any hyperparameter, based on real-time observations representing the current state of the VO system (for example, the number of points currently tracked and the distance from the past keyframe).
\blue{
Thereby minimizing the reliance on heuristic design choices while increasing the generalizability across diverse scenes and motions.
}
To enforce an accurate and robust pipeline, the reward function is constructed from the pose error between predicted and ground truth poses but also includes other non-differentiable metrics such as runtime and size of the map.
Importantly, our proposed framework leaves the underlying VO method unaltered while introducing a learned agent to enhance its robustness.
In our RL approach, we consider the underlying image sequence and the VO system as the environment.
From the environment, the agent receives observations in the form of tracked keypoints, map statistics, previously selected keyframes, and previously estimated poses.
The agent itself consists of a multi-head attention layer to downproject a variable number of keypoints, which are then further processed along the fixed-sized inputs in a two-layer MLP.
%
For training, we employ the on-policy RL algorithm Proximal Policy Optimization (PPO)~\cite{schulman2017proximal} with a privileged critic network to stabilize the training.
Finally, the reward includes the pose error between the predicted and ground truth pose computed in a sliding window approach while also penalizing decisions that lead to increased runtime.
Our experiments with the state-of-the-art VO methods~\cite{Forster17troSVO, Engel17pami} on the VO benchmarks~\cite{sturm12iros, Burri16ijrr, Geiger2012CVPR} show that our learned agent improves accuracy up to 19\%, see RL DSO in Tab.~\ref{tab:euroc_results}, and increases robustness, see RL SVO in Tab.~\ref{tab:euroc_results}, where it tracks all the camera trajectories while the w/o RL baseline gets lost.
Our contributions can be summarized as follows.

\begin{enumerate}
    \item We propose to approach the Visual Odometry (VO) problem from a sequential decision perspective, where current actions influence future states.
\blue{
    \item We introduce an RL training framework designed to train an adaptive agent to replace heuristic design choices in VO methods.     
          Our experimental results demonstrate that our approach not only boosts VO robustness and accuracy but also highlights the potential to eliminate labor-intensive fine-tuning.
}
\end{enumerate}

\section{Related Work}
\label{sec:related_work}

Three paradigms exist to approach Visual Odometry: classical geometric-based, end-to-end, and hybrid approaches.

Classical approaches have been extensively studied in computer vision and robotics in the past 30 years~\cite{scaramuzza2011visual, cadena2016past}. 
They achieve commercial-level accuracy and robustness. 
Indeed, classical VO algorithms are present in products like smartphones, VR-AR devices, and mobile robots.
Classical VO algorithms are composed of two main modules: tracking and mapping.
The tracking module uses camera images to estimate the motion.
Three main approaches exist in the literature: direct methods~\cite{Engel17pami}, feature-based methods~\cite{mur2015orb}, and semi-direct methods~\cite{Forster17troSVO}.
Direct methods, such as DSO~\cite{Engel17pami}, work directly on the raw pixel intensities.
These methods commonly extract image patches and estimate the camera trajectory by tracking the motion of such patches through consecutive images.
The tracking is achieved by minimizing a photometric error defined on the raw pixel intensities.
On the contrary, feature-based methods, such as ORB-SLAM~\cite{mur2015orb}, extract points of interest, commonly known as visual features or keypoints, from the raw image pixels.
The camera trajectory is estimated by tracking these points through consecutive images.
Feature-based methods rely on scene texture to detect and track distinctive points, while direct methods use raw image pixels and can achieve higher reliability in low-texture environments.
Semi-direct methods, such as SVO~\cite{Forster17troSVO}, combine keypoints and patches of raw pixel intensities to estimate the camera motion.
SVO uses keypoints to find regions of interest in the current camera view and raw pixel intensities to track the patches centered at the detected keypoints.

Following the deep learning revolution seen in computer vision in the past years, learning-based end-to-end VO systems~\cite{wang2017deepvo,wang2021tartanvo,ye2023pvo} have been proposed.
These methods usually rely on convolutional layers to process the input images and subsequently, recurrent and/or feedforward layers to predict the camera 6-DoF pose.
They are trained with ground truth pose supervision.
End-to-end VO systems show better performance than classical approaches in scenarios covered in the training data.
However, they lack robustness when deployed on environments (scene and motion) outside the training data distribution.

Hybrid methods~\cite{teed2021droid, teed2022deep} combine the best of two paradigms: the geometric representation of the 3D world (classical methods), and the ability to process highly dimensional inputs (end-to-end methods).
One of the state-of-the-art hybrid approaches is DROID-SLAM~\cite{teed2021droid}.
DROID-SLAM is a SLAM system that focuses on dense scene reconstruction, as well as, camera pose estimation.
It uses a RAFT-inspired network architecture~\cite{teed2020raft} followed by a differentiable dense bundle adjustment (BA) layer to iteratively update the camera poses and depth.
The BA layer includes geometry constraints that improve the consistency of consecutive camera pose estimates.
Using a similar differential BA layer, the hybrid method DPVO~\cite{teed2022deep} tracks image patches over time using a recurrent network.

The performance of VO algorithms highly depends on heuristics and hyperparameters.
Usually, hyperparameter tuning is done manually by expert users.
Due to the high dimensional space and the sequential nature of the VO problem, automatic optimization-based hyperparameters tuning methods, such as Bayesian Optimization are difficult to use.
An attempt to automatically select keyframes has been made in~\cite{sheng2019unsupervised}, where a deep network is used to decide if a new keyframe needs to be selected, as well as, to predict the camera pose.

In this paper, we model VO as a sequential decision-making task and apply Reinforcement Learning. 
Related to the problem of hyperparameter tuning with RL, Hyp-RL~\cite{jomaa2019hyp}, poses the search for the best hyperparameters for training neural networks as a sequential decision process and uses RL to select the next hyperparameter set to be evaluated.
In~\cite{dong2019dynamical}, RL is applied to dynamically choose the best set of hyperparameters of a deep network that performs object tracking.

\section{Method}
\label{sec:method}
\begin{figure*}[ht!]
\centering
\includegraphics[width=0.95\textwidth]{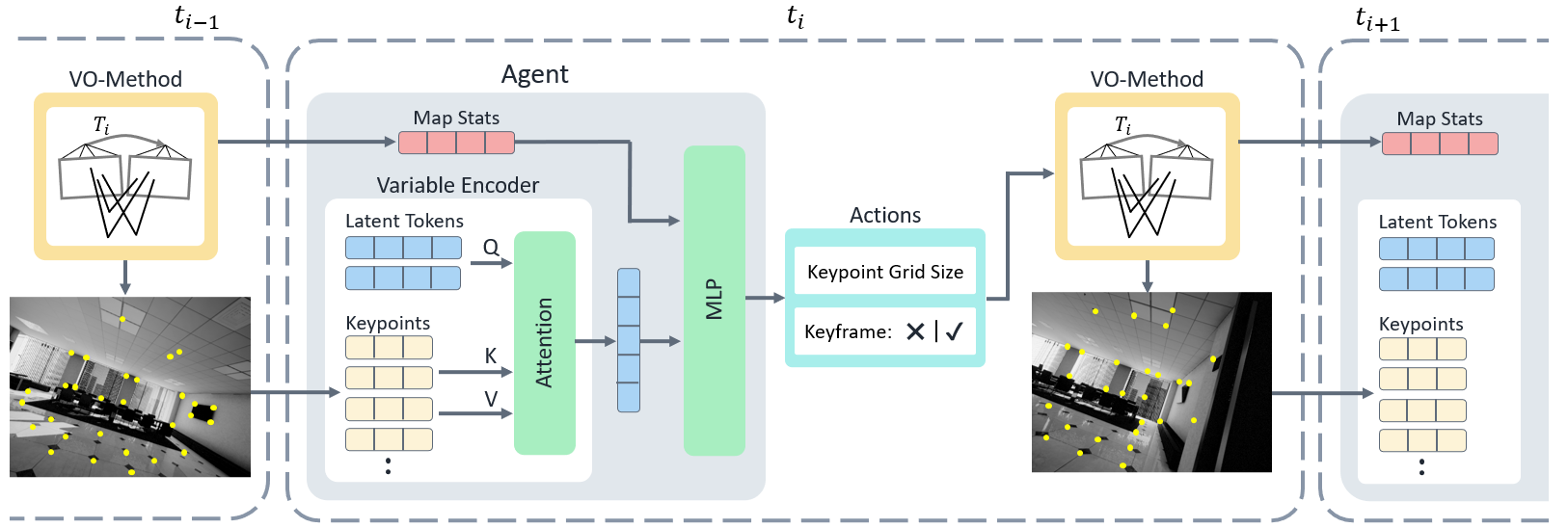}
\caption{
\textbf{Agent Overview.}
Our agent takes as input the map statistics computed by the VO method at timestep $t_{i-1}$ and a variable number of keypoints using a multi-head attention layer.
Given these inputs, a two-layer MLP computes a multi-discrete probability distribution over the binary keyframe action and the grid size action. 
These actions are then used inside the VO method to process the current frame at timestep $t_i$, which will lead to the observation for the next timestep $t_{i+1}$.
}
\label{fig:agent_overviwe}
\end{figure*} 
We consider monocular Visual Odometry (VO), whose primary objective is to estimate the camera pose for each image within an image sequence.
Despite extensive research and widespread application, the robustness and generalizability of VO methods in different scenes remain an ongoing challenge.
Both traditional geometry-based approaches and hybrid methods incorporating learned components often rely on heuristic decision rules for critical tasks such as keyframe selection and keypoint filtering.
These heuristic rules significantly affect the performance of the VO method and are closely linked to the deployed scene.
Consequently, a substantial amount of fine-tuning is required to adapt these rules effectively to each deployed scenario.
To increase the robustness and generalizability of VO pipelines, we formulate the VO task as a sequential decision problem.
Within this sequential framework, a deep agent is trained through RL.
This agent is designed to make decisions at each timestep for the VO method, taking into account the current state of the system, including map statistics, keypoints information and previous pose estimations.
By training an adaptive agent, we can effectively reduce the reliance on heuristics and increase the generalizability of the underlying VO method across diverse scenes.
The structure of this section is the following. In Sec.~\ref{sec:problem_formulation}, we present our problem formulation. 
In Sec.~\ref{sec:agent}, we introduce our proposed deep neural agent with its corresponding input observations and output actions.
In Sec.~\ref{sec:reward}, we formulate the reward function, while in Sec.~\ref{sec:training}, we present the RL training framework. 
Finally, in Sec.~\ref{sec:vo_methods}, we explain the specific implementation of our RL framework using SVO~\cite{Forster17troSVO} and DSO~\cite{Engel17pami} as the underlying VO methods.

\subsection{Problem Formulation}
\label{sec:problem_formulation}
We formulate the VO task as a sequential decision problem in which an agent interacts with its environment, see Fig.~\ref{fig:method_overview}.
Specifically, we consider the image sequence and the underlying VO pipeline to be fixed, both constituting the \emph{environment}.
The agent is a neural network that, at each timestep, selects \emph{actions} for the VO pipeline, i.e., selecting keyframes and adapting the keypoint distribution by setting the grid size constraint (a grid is fitted to the image, and maximum one keypoint per grid cell is selected).
Based on these actions, the VO method processes the next image of the sequence and computes the corresponding pose.
With this definition of the agent and environment, we formulate the VO task as a Markov decision process (MDP), denoted by the tuple $M= (S, A, p, r, \gamma)$.
A state $s \in S$ encapsulates the current image in the image sequence and the state of the VO pipeline, which includes properties such as feature tracks, previously estimated poses, the triangulate map, the optimization state, etc.
To transition between VO states, actions $a \in A$ are executed in the VO method, while the environment supplies the next frame of the sequence. 
The state transition probability $p$ defines the likelihood of moving from one state $s_i$ to another $s_j$ given actions $a_i$.
Finally, each state transition is associated with a reward $r(s, a)$, dependent on the ensuing state and the chosen actions.
The rewards of the future are discounted by $\gamma$.
In the context of VO, the reward function evaluates the accuracy of the pose estimation and can also incorporate non-differentiable criteria such as runtime and map size, among others.
In our formulation, the agent is characterized by a control policy $\pi: S \times A \rightarrow \mathbb{R}$, where $\pi$ specifies the actions $a \in A$ to be taken in each state $s \in S$.
Since the control policy is non-deterministic, it induces a probabilistic trajectory of states and actions $\tau = \{s_0, a_0, s_1, a_1, ..., s_T, a_{T-1}\}$, abbreviated as $\tau \sim \pi$.
Following the control policy $\pi$, the expected sum over the discounted rewards for each timestep can be expressed with its value function
\begin{equation}
    V^{\pi}(s_i) = \mathbb{E}_{\tau \sim \pi} \left[ \sum_{t=i}^{\infty} \gamma^t r(s_t, a_t) \right].
\label{eq:expected_return}
\end{equation}
The overarching goal is to find a policy $\pi$ that maximizes the expected sum of discounted rewards.
\begin{equation}
\label{eq:opt_problem}
    \pi^* = \arg\max_{\pi}  V^\pi(s_0)
\end{equation} 
Given the reward formulation, the optimal policy provides decisions to the VO pipeline, aiming to minimize the expected position error while accounting for additional reward terms like runtime.

\subsection{Deep Neural Agent}
\label{sec:agent}
The central element of our proposed framework is the learned agent embodying the policy $\pi$, which selects optimal settings for the VO method based on the current keypoints and map statistics, i.e., relative poses to the newest and oldest keyframe.
Given the inherent complexity of the underlying VO method, supplying the agent network with a comprehensive state encompassing all stored map elements, optimization states, etc., is neither feasible nor advantageous.
Instead, the learned agent receives a selected subset of observations $o \in  O$ approximating the VO state within the policy $\pi: O \times A \rightarrow \mathbb{R}$.
In our system, the observations are derived from the local map and include relative poses to previous keyframes, along with the information about the currently tracked keypoints.
Notably, the number of keypoints is variable and may vary depending on the characteristics of the image sequence.
Therefore, the network architecture of the agent should be capable of processing inputs of variable sizes.

To handle inputs of varying sizes, we leverage a Perceiver~\cite{jaegle2021perceiver, jaegle2022perceiverio} architecture, illustrated in Fig.~\ref{fig:agent_overviwe}.
Specifically, our approach involves a \textit{Variable Encoder} that incorporates a multi-head attention layer~\cite{Vaswani17neurips} for processing the image positions, and the estimated depth of each tracked keypoint $k \in \mathbb{R}^{N\times 3}$.
To project the information contained in the variable number of keypoints $N$ to a fixed dimension, \camready{we use $M$ number of learned tokens $z \in \mathbb{R}^{M\times D}$ with dimension of $D$} as query $Q$ values.
These query tokens are attended to by the projected keypoint information serving as both key $K$ and values $V$.
%
%
Finally, the fixed output of the Variable Encoder is subsequently processed along with the statistics of the local map, using a two-layer MLP with ReLu non-linearities.
The output of the agent network is a discrete probability distribution over a set of actions expressed in the form of multiple independent categorical distributions.
In our implementation, the agent predicts at each timestep a binary distribution over the decision to take the current frame as a keyframe.
%
%
Additionally, in the case of SVO, the agent also predicts a separate and independent categorical distribution over a set of values $\{20, 25, ..., 40\}$ for the keypoint grid size.
This grid size ensures a uniform distribution of keypoints on the image plane by enforcing that only one keypoint can reside within a given grid cell.
Since DSO does not extract keypoints, there is no grid size selection necessary.

\subsection{Reward}
\label{sec:reward}
\begin{figure}[ht!]
\centering
\includegraphics[width=0.7\textwidth]{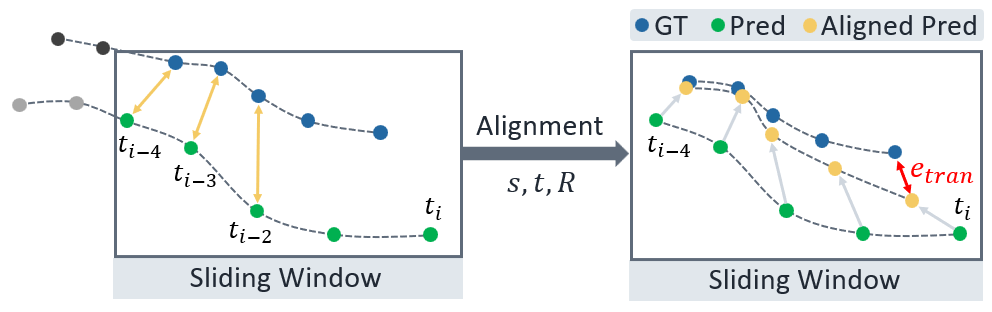}
\caption{
\textbf{Position Error.}
%
%
To closely relate the pose prediction accuracy to the current action, we employ a sliding window of five timesteps to align the ground truth and estimated trajectory using a scale factor $s$, a translation vector $t$, and a rotation $R$.
The error at the current timestep $t_i$ is then used as the negative position reward.
}
\label{fig:position_error}
\end{figure} 
Another critical component within the RL framework is the design of the reward function.
The feedback of the reward at each timestep guides the deep agent toward a desired behavioral pattern.
In the case of VO, naturally, the error between the estimated and the ground truth pose is the primary focus of the reward.
However, since monocular VO methods cannot inherently estimate the absolute scale of the trajectory, it becomes necessary to scale the predicted poses before computing the error against the ground truth trajectory.
Furthermore, another crucial aspect of ensuring stable feedback for the RL algorithm is to closely link the reward function of a timestep $t_i$ to the action taken at that specific timestep, thus mitigating the value assignment problem.
Consequently, we adopt a sliding window approach to compute the error between the estimated and ground truth poses.
Specifically, we employ a window of five timestamps $\{t_{i-4}, t_{i-3}, ..., t_{i}\}$ to compute the error at the current timestep $t_i$, see Fig.~\ref{fig:position_error}.
The first three poses in the window are used to align the estimated trajectory to the ground truth by minimizing the distances between the positions with the Umeyama method~\cite{Umeyama91pami}.
This alignment method accommodates missing scale information and differing coordinate origins, aligning both trajectories through an Euclidean transformation involving rotation, translation, and scaling.
By designing the position reward function, we can influence the policy to weigh accuracy higher at the cost of lower robustness or the other way around.
This trade-off also shows the benefit of our dynamic agent over manual tuning since the decisions can be made in an online fashion, adapting to the current conditions in the image sequence.
To handle this trade-off, we clip the positional error $e_{\text{tran}}$ around zero, see Eq.~\ref{eq:reward}, to positively enforce accurate states while giving a penalty for states having a large positional error.
Given that our framework leverages RL, we have the opportunity to integrate non-differentiable performance feedback into the reward system. 
Thus, we can enforce a faster runtime by penalizing the insertion of keyframes, i.e., directly penalizing the keyframe action $a_{\text{keyframe}}$.
Using the keyframe action instead of directly penalizing the runtime mitigates the dependence on the usage of the hardware during training.
Our final reward function is formulated in Eq.~\ref{eq:reward}, where we use 0.01 for $\lambda_1$ and set $\lambda_2$ to \num{5e-3}.
\begin{align}
\begin{split}
    r =& \lambda_1 \max(-1, 0.2 - e_{\text{tran}}) - \lambda_2 a_{\text{keyframe}}
\end{split}
\label{eq:reward}
\end{align}

\subsection{Reinforcement Learning Framework}
\label{sec:training}
%
We use the on-policy algorithm Proximal Policy Optimization (PPO)~\cite{schulman2017proximal} implemented in Stable Baselines3~\cite{Raffin21jmlr} for training our agent.
As commonly done, the RL training is conducted in two alternating stages: the rollout and the policy-update stage.
During the rollout stage, the RL agent interacts with the environment, i.e., the underlying VO method and image sequences from TartanAir~\cite{wang20iros}, via its actions to collect experiences comprising the actions, observations, and rewards at each timestep.
While the experiences are collected, the network weights of the agent are not updated.
In the second stage, the collected experiences of the rollout stage are used to update both the agent and the critic.
\camready{
The task of the critic is to approximate the value function, which represents the expected return of the current policy at a given state.
The critic is used to stabilize the training inside the policy update steps in accordance with the PPO algorithm, which is designed to limit large changes in the policy.
}

Since the performance of the VO pipelines depends not only on the decisions of the agent but also on the current image sequence, we introduce a privileged critic to stabilize the training.
This privileged critic has access to both the current and future ground truth poses, enabling a better assessment of whether tracking performance was influenced by the decisions of the agent or the difficulty of the underlying image sequence. 
Importantly, this privileged critic with access to ground truth poses is only used during training.
Another crucial aspect is the handling of the different stages of the VO system, which include the normal mode of camera tracking, the mode for the relocalization in the current map, and the initialization mode.
Since the VO system only provides information about its full state in the tracking mode, we declare the current state a valid state if the previous state was in the tracking mode.
Notably, the agent can exclusively make decisions in valid stages.
To account for the valid states in the policy-update phase, we use a masked rollout buffer, which computes the return using all collected states, but samples only valid states for updating the policy network.
As states in the initialization stages can not be influenced by the agent, we directly use the ground truth poses to triangulate tracked keypoints.
%
The initialization with the ground truth poses also has the benefit of a significantly faster runtime compared to the original initialization.
Since RL is sample inefficient, it is crucial to have a fast runtime of the environment interactions.
For our RL framework, the runtime of each of the main components in the rollout phase using 100 SVO instances in parallel is reported in Tab.~\ref{tab:training_runtime}.

\subsection{VO System}
\label{sec:vo_methods}
In general, our proposed agent network can be combined with any VO system that relies on decision components.
In this work, we use SVO~\cite{Forster17troSVO} and DSO~\cite{Engel17pami}, which feature a very small runtime.
Additionally, we also test on the slower SLAM pipeline ORB-SLAM3~\cite{ORBSLAM3_TRO}.
%
%
From a practical viewpoint, the small runtime and parallelization accelerate training and partially compensate for the inherent low sample efficiency of RL.

SVO is a semi-direct visual odometry composed of two main threads: motion estimation and mapping thread.
The first step of the motion estimation thread is \textit{image alignment}, where the pose of the new frame is estimated by aligning pixel intensity patches from the previous keyframe.
The next step is \textit{feature alignment}, where the positions of the keypoints on the current image are refined while the camera poses are fixed.
The last step is a sparse bundle adjustment, where both the camera poses and the locations of the 3D points included in the current map are optimized.
The motion thread assumes that the depth of the mapped 3D points is known. 
It is the objective of the mapping thread to estimate the depth.
The mapping thread estimates the depth of each tracked pixel from multiple observations by means of a recursive Bayesian filter~\cite{vogiatzis2011video}.

DSO is a direct visual odometry also composed of two threads: the front-end thread and the mapping thread. 
The front-end thread performs the initial frame tracking where the pose of the current frame is tracked with respect to the previous keyframe by aligning pixel intensities patches. The front-end is also in charge of selecting the active candidate points, performing outlier rejection and occlusion detection, and selecting keyframes.
The mapping thread optimizes a non-linear cost function based on the \textit{photometric error}. The camera poses, intrinsics, and inverse depth value of the active points in the current map are jointly optimized.
This optimization can be seen as the photometric equivalent of sparse bundle adjustment.

We use SVO to run ablation studies on the contributions of the main components of the RL system (runtime, inputs, actions, and reward tuning).
\newcolumntype{C}[1]{>{\centering}m{#1}}

\section{Experiments}
\label{sec:experiments}

\noindent \textbf{Dataset}
For training our agent, we leverage the synthetic sequences from TartanAir~\cite{wang20iros}, a comprehensive dataset offering, among other measurements, ground truth camera poses.
The dataset features a wide range of simulated scenes with challenging conditions such as low light and weather effects.
The total number of different training sequences is 337, amounting to 279987 images or timesteps for the training of the agent.
Importantly, we consider the dataset in our RL framework to be fixed without the possibility of the agent influencing the next image. 
Notably, as our method is trained on abstractions rather than image pixels, the risk of domain shift is mitigated.
For evaluation, we use three VO benchmarks: EuRoC~\cite{Burri16ijrr}, TUM-RGBD~\cite{sturm12iros} and KITTI~\cite{Geiger2012CVPR}.
The EuRoC dataset consists of 11 sequences recorded via a drone, with ground truth pose measurements obtained from a motion capture system or a laser tracker.
The TUM-RGBD dataset features 9 sequences recorded via a hand-held camera with ground truth poses provided by a motion capture system.
%
%
For the results on the KITTI dataset, we refer to the supplementary.
\noindent \textbf{Evaluation}
%
%
Following standard practice in evaluating VO algorithms, we use the Absolute Translation Error (ATE [m])~\cite{zhang2018tutorial} as the error metric for the EuRoC and TUM-RGBD datasets. 
%
%
Before computing the error metric, the estimated trajectories are aligned to the ground truth trajectory using the Umeyama method~\cite{Umeyama91pami}.
For the relative position error on EuRoC and TUM-RGBD, we refer to the supplementary.

Since we want to guarantee a fair comparison between the methods with and without RL, we remove the effect of the initialization phase at the beginning of each sequence by providing the ground truth poses to the VO algorithm.
Similar to the original initialization of the underlying VO, extracted keypoints are tracked from the first frame of the sequence and triangulated once a large enough disparity is detected.
However, instead of estimating a relative pose, the ground truth poses are used for the triangulation.
Following this procedure, all of the tested methods start with the same high-quality map, effectively eliminating the unwanted influence of the initialization.
Naturally, the ground truth initialization is not used for the comparison against state-of-the-art VO methods. 
Instead, we take the average of the performance metrics for five runs.
\begin{figure}[ht!]
\begin{minipage}{.475\linewidth}

\centering
\includegraphics[width=1\textwidth]{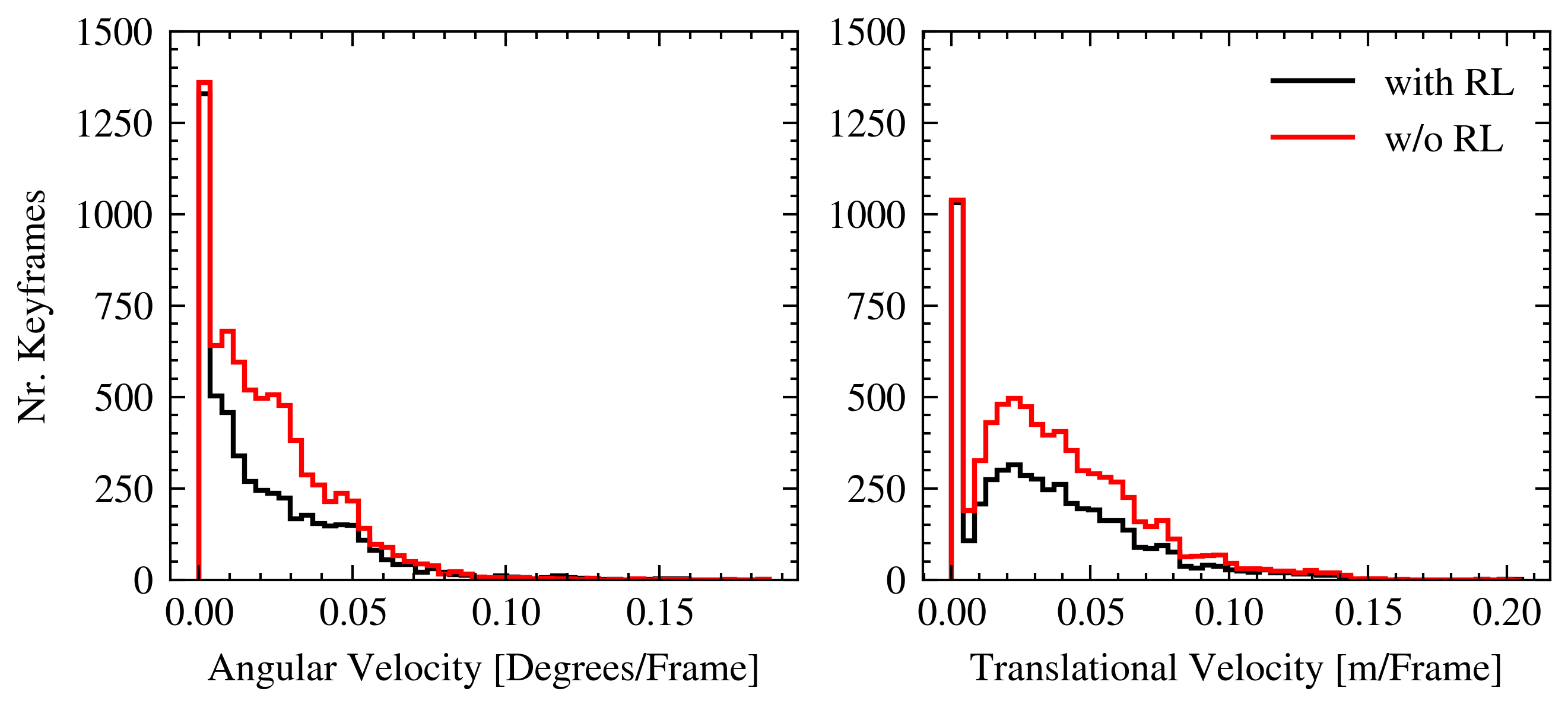}
\caption{
\textbf{Keyframe Selection.}
The number of selected keyframes by SVO at different translational and angular velocities for the EuRoC dataset.
}
\label{fig:keyframe}

\end{minipage}
\hfill
\begin{minipage}{.475\linewidth}

\centering
\includegraphics[width=0.99\textwidth]{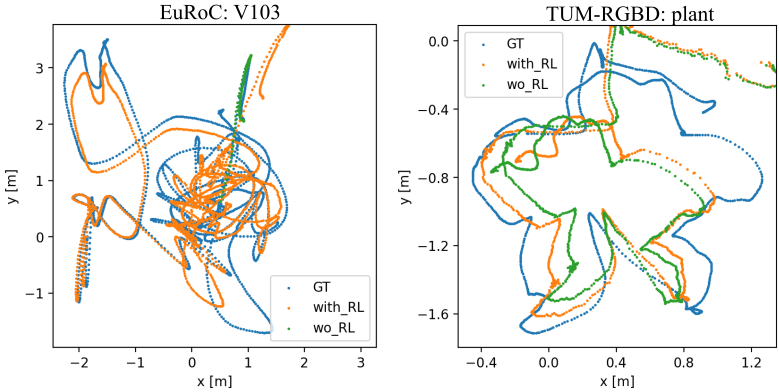}
\caption{
\textbf{Top-Down View of Trajectories.}
Visualization of the predicted trajectory using SVO with and without RL.
}
\label{fig:top_down_red}

\end{minipage}
\end{figure}
\begin{table}[ht!]

\newcommand{\cw}{1.1cm}
\centering
\scalebox{0.7}{
\begin{tabular}{m{2.7cm}C{\cw}C{\cw}C{\cw}C{\cw}C{\cw}C{\cw}C{\cw}C{\cw}C{\cw}C{\cw}C{\cw}>{\centering\arraybackslash}m{\cw}}
Method  & MH01 & MH02 & MH03 & MH04 & MH05 & V101 & V102 & V103 & V201 & V202 & V203 & Avg \\
\hline
SVO     & \textbf{0.183}         & 0.490          &          0.381 & 3.236          & \textbf{1.293} & 0.320          & \textbf{0.333} &  -    & 0.084 & 0.553 & \textbf{1.431} & -\\
RL SVO          & 0.236          & \textbf{0.124} & \textbf{0.241} & \textbf{2.149} & 1.982          & \textbf{0.225} & 0.398 & \textbf{0.394} & \textbf{0.069} & \textbf{0.334} & 1.506 & \textbf{0.696} \\
\hline
DSO             & 0.087          & 0.044          &         0.354  & 0.184          & \textbf{0.260} & 1.223          & 1.624 & \textbf{1.125}  & 0.059 & \textbf{0.065} & 1.516 & 0.594 \\
RL DSO          & \textbf{0.037} & \textbf{0.033} & \textbf{0.241} & \textbf{0.168} & 1.192 & \textbf{0.207} & \textbf{0.499} & 1.450  & \textbf{0.043} & 0.164 & \textbf{1.286} & \textbf{0.483} \\
\end{tabular}}
\caption{
\textbf{EuRoC.}
We report the ATE [m] of our RL agents and the heuristic-based VO methods with parameters obtained by an extensive grid search.
Our RL agents outperform the heuristic-based VO methods in terms of accuracy (lower average error) and robustness (RL SVO can complete the sequence \textit{V103} while the baseline SVO fails to track all the camera poses due to fast rotations).}
\label{tab:euroc_results}
\end{table}
\begin{table*}[ht!]

\newcommand{\cw}{1.1cm}
\centering
\scalebox{0.75}{
\begin{tabular}{m{2.9cm}C{\cw}C{\cw}C{\cw}C{\cw}C{\cw}C{\cw}C{\cw}C{\cw}C{\cw}>{\centering\arraybackslash}m{\cw}}

Method  & 360 & desk & desk2 & floor & plant & room & rpy & teddy & xyz & Avg  \\
\hline
SVO & \textbf{0.186} & 0.681 & 0.898 & - & 0.320 & 0.805 & \textbf{0.053} & 0.769 & \textbf{0.057} & 0.471 \\
RL SVO          & 0.189 & \textbf{0.556} & \textbf{0.755} & - & \textbf{0.260} & \textbf{0.804} & 0.056 & \textbf{0.697} & 0.062 & \textbf{0.422}  \\
\hline
DSO             & -              & \textbf{0.237} & 0.893          & - & \textbf{0.320} & \textbf{0.931} & 0.057          & -      & \textbf{0.053} & - \\
RL DSO          & \textbf{0.187} & 0.682          & \textbf{0.888} & - & 0.625          & -              & \textbf{0.054} & \textbf{0.756} & 0.105  & - \\

\end{tabular}}
\caption{
\textbf{TUM-RGBD.}
We report the ATE [m] of our RL agents and the heuristic-base VO methods with parameters obtained by an extensive grid search.
Our RL SVO outperforms the heuristic-based SVO methods in terms of accuracy (lower average error).
Although this dataset is challenging for DSO, due to the use of a rolling shutter camera, RL DSO improves the robustness with two more sequences completed.
}
\label{tab:tum_results}
\end{table*}
\begin{table*}[ht]

\scalebox{0.7}{
\newcommand{\cw}{1.1cm}
\centering
\begin{tabular}{m{2.5cm}C{\cw}C{\cw}C{\cw}C{\cw}C{\cw}C{\cw}C{\cw}C{\cw}C{\cw}C{\cw}C{\cw}>{\centering\arraybackslash}m{\cw}}
 Method  & MH01 & MH02 & MH03 & MH04 & MH05 & V101 & V102 & V103 & V201 & V202 & V203 & Avg  \\
\hline
TartanVO ~\cite{wang2021tartanvo} & 0.639          & 0.325          & 0.550          & 1.153          & 1.021          & 0.447          & 0.389 & 0.622  & 0.433 & 0.749 & 1.152 & 0.680 \\
DROID-VO ~\cite{teed2021droid}    & 0.163          & 0.121          & 0.242          & 0.399          & 0.270          & 0.103          & 0.165          & 0.158 & 0.102 & 0.115 & \textbf{0.204} & 0.186 \\
DPVO ~\cite{teed2022deep}         & 0.087          & 0.055          & \textbf{0.158} & \textbf{0.137} & 0.114 & \textbf{0.050} & 0.140 & \textbf{0.086} & 0.057 & \textbf{0.049} & 0.211 & \textbf{0.105} \\
RL SVO                            & 0.289 & 1.599 & 0.552 & 2.407   & 1.826 & 0.229 & 0.813 & 0.496 & 0.085 & 0.965 & 1.434 & 0.972 \\
RL DSO                            & \textbf{0.032} & \textbf{0.037} & 0.166 & 0.142 & \textbf{0.110} & 0.107 & \textbf{0.089} & 1.331 & \textbf{0.041} & 0.152 & 1.313 & 0.320
\end{tabular}}
\caption{
\textbf{EuRoC.}
We report the ATE [m] of our RL agents and the state-of-the-art VO methods. Our RL DSO achieves the best accuracy in 5 sequences.}
\label{tab:sup_euroc_results}
\end{table*}
\begin{table*}[ht]

\newcommand{\cw}{1.1cm}
\centering
\scalebox{0.75}{
\begin{tabular}{m{2.9cm}C{\cw}C{\cw}C{\cw}C{\cw}C{\cw}C{\cw}C{\cw}C{\cw}C{\cw}>{\centering\arraybackslash}m{\cw}}

 Method  & 360 & desk & desk2 & floor & plant & room & rpy & teddy & xyz & Avg  \\
\hline

DROID-VO ~\cite{teed2021droid}    & 0.161          & \textbf{0.028 }         & 0.099          & \textbf{0.033} & \textbf{0.028} & \textbf{0.327} & \textbf{0.028} & 0.169          & 0.013 & 0.098 \\
DPVO ~\cite{teed2022deep}         & \textbf{0.135} & 0.038          & \textbf{0.048} & 0.040          & 0.036          & 0.394          & 0.034          & \textbf{0.064} & \textbf{0.012} & \textbf{0.089} \\
RL SVO           & 0.183 & 0.648 & 0.790 & -     & 0.332 & 0.786 & 0.052 & 0.741 & 0.071 & - \\
RL DSO           & -     & 0.438 & 0.765 & -     & 0.479 & -     & -     & -     & 0.057 & -
\end{tabular}}
\caption{
\textbf{TUM-RGBD.}
We report the ATE [m] of our RL agents and the state-of-the-art VO methods. }
\label{tab:sup_tum_results}
\end{table*}
\begin{table}[ht!]

\newcommand{\cw}{1.1cm}
\centering
\scalebox{0.65}{
\begin{tabular}{m{3.9cm}C{\cw}C{\cw}C{\cw}C{\cw}C{\cw}C{\cw}C{\cw}C{\cw}C{\cw}C{\cw}C{\cw}>{\centering\arraybackslash}m{\cw}}

 Method  & MH01 & MH02 & MH03 & MH04 & MH05 & V101 & V102 & V103 & V201 & V202 & V203 & Avg  \\
  
\hline
SVO           & 0.183 & 0.490 & 0.381 & 3.236 & 1.293 & 0.320 & 0.333 &  -    & 0.084 & 0.553 & 1.431 & - \\
RL SVO                & 0.236 & 0.124 & \textbf{0.241} & 2.149 & 1.982 & 0.225 & 0.398 & \textbf{0.394} & \textbf{0.069} & 0.334 & 1.506 & \textbf{0.696} \\
\hline
RL SVO (w/o keypoints)       & 0.255 & 0.160 & 0.419 & \textbf{1.979} & \textbf{0.588} & \textbf{0.173} & 0.644 & 0.849 & 0.182 & 1.394 & - & -\\
RL SVO (w/o privileged c.)   & \textbf{0.130} & \textbf{0.099} & 0.834 & 3.283 & 0.958 & 0.140 & 0.286  & 0.459 & 0.116 & 0.618 & \textbf{1.373} & 0.754 \\
\hline
RL SVO (w/o keyframe) & 0.179 & 0.351 & 0.302 & 3.072 & 1.021 & 0.325 & -     & -      & 0.179 & \textbf{0.258} & 1.464 & - \\
RL SVO (w/o gridsize) & 0.152 & 0.183 & 0.293 & 2.154 & 1.537 & 0.198 & \textbf{0.269} & 0.808 & 0.108 & 1.683  & 1.405 & 0.799 \\
\end{tabular}}
\caption{
\textbf{EuRoC Ablations.}
Ablation study on inputs and action of the RL agent.
}
\label{tab:euroc_ablation}
\end{table}
\begin{table}[ht!]
\begin{minipage}{.475\linewidth}
\centering
\scalebox{0.75}{
\begin{tabular}{m{3.9cm}>{\centering\arraybackslash}m{2.8cm}}
Components  & Time [ms]\\
\hline
- 100 Parallel SVO Step    &  73.9 \\
- Load 100 Images          &  18.6 \\
- Network Forward Pass     &   3.1 \\
Complete Rollout Step      & 104.1 \\
\end{tabular}}
\vspace{0.35cm}
\caption{
\textbf{Rollout Runtime.}
The runtime for the main components in the rollout phase averaged over 500 iterations.
}
\label{tab:training_runtime}
\end{minipage}
\hfill
\begin{minipage}{.475\linewidth}
\centering
\scalebox{0.75}{
\begin{tabular}{m{3.9cm}>{\centering\arraybackslash}m{2.8cm}}
Method  & Time [ms]\\

\hline
SVO                  &  9.06 \\
RL SVO               &  7.40 \\
\hline
DSO                  &  34.45 \\
RL DSO               &  32.29 \\
\hline
Network inference    &  1.75 \\
\end{tabular}}
\caption{
\textbf{Inference Runtime.}
The runtime of SVO, DSO, and one network forward pass.
}
\label{tab:runtime}
\end{minipage}
\end{table}
\begin{table*}[!b]
\newcommand{\cw}{1.1cm}
\centering
\scalebox{0.7}{
\begin{tabular}{m{2.7cm}C{\cw}C{\cw}C{\cw}C{\cw}C{\cw}C{\cw}C{\cw}C{\cw}C{\cw}C{\cw}C{\cw}>{\centering\arraybackslash}m{\cw}}
 & \multicolumn{12}{c}{EuRoC (ATE [m])} \\
 \cmidrule(lr){2-13}
Method  & MH01 & MH02 & MH03 & MH04 & MH05 & V101 & V102 & V103 & V201 & V202 & V203 & Avg  \\  
\hline
ORB-SLAM3             & 0.031 & 0.032  & 0.041 & 1.256          & \textbf{0.066} & 0.035          & 1.528          & 1.451          & \textbf{0.201} & 0.190 & \textbf{1.469} & 0.573 \\
RL ORB-SLAM3          & \textbf{0.029} & \textbf{0.028} & \textbf{0.039} & \textbf{0.152} & 0.106 & \textbf{0.034} & \textbf{1.472} & \textbf{1.441} & 0.203 & \textbf{0.070} & 1.654 & \textbf{0.475} \\
%
 & \multicolumn{12}{c}{TUM-RGBD (ATE [m])} \\
\cmidrule(lr){2-11}
 Method             & 360   & desk  & desk2 & floor & plant & room  & rpy   & teddy & xyz   & Avg   &  & \\
\cmidrule(lr){1-11}
ORB-SLAM3           & -     & \textbf{0.031} & 0.746          & \textbf{0.013} & 0.154          & 1.048          & - & 0.108          & \textbf{0.011} & 0.302          & & \\
RL ORB-SLAM3        & -     & \textbf{0.031} & \textbf{0.604} & \textbf{0.013} & \textbf{0.038} & \textbf{0.717} & - & \textbf{0.091} & \textbf{0.011} & \textbf{0.215} & & \\
\end{tabular}}
\caption{
\textbf{ORB-SLAM3.}
We report the ATE [m] of our RL agent applied on ORB-SLAM3.
}
\label{tab:orbslam_results}
\end{table*}

\subsection{Results} \label{sec:results}

\noindent \textbf{EuRoC}
We compare our RL agent with SVO and DSO against the heuristic-based VO with parameters obtained by tuning on this dataset using an extensive grid search.
Qualitative estimation results of SVO are shown in the left plot in Fig.~\ref{fig:top_down_red} from a top-down view of the trajectory estimated by RL SVO, the baseline VO, and the ground truth.
Tab.~\ref{tab:euroc_results} reports the ATE.
In the case of SVO, the combination with our RL agent achieves the best ATE in most of the sequences.
Remarkably, our RL SVO can estimate the entire camera trajectory in the sequence \textit{V103} while the baseline fails.
This sequence is one of the most difficult in the EuRoC dataset due to fast camera rotations.
The benefit of our RL agent is also evident for DSO, which achieves consistently a lower tracking error, leading to a 19\% lower mean tracking error. 
We report in Tab.~\ref{tab:sup_euroc_results} the comparison of our RL agent against the state-of-the-art VO methods, which are taken from~\cite{teed2022deep}.
The results reported for our methods are averaged over five runs.
Our RL DSO achieves the best ATE in five sequences.
\noindent \textbf{TUM-RGBD}
Similar to EuRoC, we use a parameter grid search to find the best parameters for the baseline DSO and SVO methods.
We report the ATE in Tab.~\ref{tab:tum_results}, which shows that SVO RL achieves a lower ATE in most of the sequences and a lower overall mean error.
Since the images of TUM-RGBD are not captured with a global shutter camera, DSO, relying on photometric alignment, struggles to finish all the sequences.
Nevertheless, our RL agent improves the robustness in some challenging conditions with two more sequences successfully completed.
For completeness, the top-down view of the trajectory estimated by RL SVO, the baseline, and the ground truth are visualized in the right plots in Fig.~\ref{fig:top_down_red}.
We report in Tab.~\ref{tab:sup_tum_results} the comparison of our RL agent against the state-of-the-art VO methods.
The results reported for our methods are averaged over five runs.
The results of the state-of-the-art methods are taken from~\cite{teed2022deep}.
%

%

%
\noindent \textbf{Keyframe Selection}
In Fig.~\ref{fig:keyframe}, we give insights into the keyframe selection strategy employed by our RL agent. 
We visualize the histogram of states at which keyframes are selected and their corresponding rotational and translation velocity.
The keyframe selection using the RL agent has a similar distribution around the motion patterns, while the heuristic rules select more keyframes.
\camready{
Following the motion distribution of the dataset, both methods trigger more keyframes in small motions.
}
%
%
Despite the fewer keyframes, our RL agent still achieves a more robust performance.
Without the possibility to set keyframes, namely, the agent only determines the feature grid size, robustness is drastically reduced, as it can be seen in Tab.~\ref{tab:euroc_ablation} by the entry RL SVO (w/o keyframe).
\noindent \textbf{Grid size Selection}
%
%
The positive impact of dynamically selecting the grid size for SVO can be observed by removing the grid size action, which lowers the pose estimation accuracy, see RL SVO (w/o gridsize) in Tab.~\ref{tab:euroc_ablation}.
The supplementary video shows that the RL agent learns to reduce the grid size to the minimal size of 20 for challenging conditions, leading to more keypoints being tracked.
Another behavior that can be observed is the quick increase of the grid size for one or two frames.
A possible explanation for this short increase and subsequent decrease in the grid size is that the agent tries to filter out less robust keypoints.

\noindent \textbf{Runtime}
%
%
The runtime achieved by DSO and SVO given decisions from the heuristic rules and our RL agent are reported in Tab.~\ref{tab:runtime}.
The results are obtained by averaging the runtime required for processing all sequences of EuRoC on one NVIDIA A100.
Our RL agent leads to a faster average SVO runtime, which can be related to the lower rate of keyframe selection, see Fig.~\ref{fig:keyframe}.
In comparison, RL DSO does not lead to a high runtime reduction.
%
%
For the performance of the agents with different penalty terms, we refer to the supplementary. 
Finally, the architecture of our agent is very lightweight, with a total of 296,584 parameters.
%

\noindent \textbf{SLAM Method}
In addition to VO methods, we also tested our method with the full SLAM pipeline ORB-SLAM3. 
Differently from the VO methods, which run in a sliding-window fashion (old measurements are marginalized), ORB-SLAM3 uses a local bundle adjustment that has access to the entire history of keyframes.
Although the local bundle adjustment could compensate for uninformative keyframes, our agent can still provide a more suitable keyframe selection strategy than the heuristics, as it can be observed in Tab.~\ref{tab:orbslam_results} by the increased accuracy on the EuRoC and TUM-RGBD datasets.
To run ORB-SLAM3 with our framework, significant adjustments with respect to the synchronization between the tracking and bundle adjustment threads were required, amongst others, the removal of the loop closure algorithm, which explains the different performance compared to results reported in~\cite{ORBSLAM3_TRO}.

\noindent \textbf{Ablation}
To evaluate the benefits of our \textit{Variable Encoder} processing the keypoints, we report in Tab.~\ref{tab:euroc_ablation} the performance of an agent network without the Variable Encoder, i.e., RL SVO (w/o keypoints).
As can be observed, the agent without access to the keypoint information is not capable of successfully finishing all the sequences of the EuRoC dataset.
Furthermore, we also report the performance of the agent trained without privileged critic referred to as SVO RL (w/o privileged c.), leading to a larger mean ATE.
For the ablations on TUM-RGBD, we refer to the supplementary.
%

\section{Conclusion}
By approaching the Visual Odometry (VO) problem from a sequential decision-making perspective, we employ a neural network trained within a Reinforcement Learning (RL) framework to dynamically adapt the VO method at each timestep. 
Instead of hand-tuned heuristics and fixed hyperparameters, the dynamic agent predicts optimal actions for the VO method to increase its accuracy and robustness.
The RL agent is trained on a large-scale synthetic dataset and tested on common real-world VO benchmarks showing improved robustness and accuracy.
%
%
We believe this paradigm shift advances VO technology and opens avenues for integrating RL into visual inertial odometry and simultaneous localization and mapping approaches.

\section*{Acknowledgments}
This work was supported by the European Research Council (ERC) under grant agreement No. 864042 (AGILEFLIGHT).
%
\section*{\Large \bf Supplementary: Reinforcement Learning Meets Visual Odometry}

\section{Limitations}\label{sec:limitations}
One of the main challenges for the RL framework is the difficulty of the TartanAir sequences used for training.
If the underlying VO method struggles to complete the sequences, our RL framework can experience instabilities due to wrong decision rewards.
This can affect the final performance, even with the employment of a privileged critic.
Furthermore, the trained RL agent also struggles with motion patterns not present in the training dataset, which especially includes static motions.

\section{Training Details}
\label{sec:training_details}
We train our agent network using the on-policy algorithm Proximal Policy Optimization (PPO)~\cite{schulman2017proximal} implemented in Stable Baselines3~\cite{Raffin21jmlr}.
The environment is comprised of image sequences from TartanAir~\cite{wang20iros} processed within 100 parallel instances of the chosen VO algorithm.
The sequences are selected randomly inside each environment at the start of the training and when the current sequence finishes.
The agent is trained for a total of 1000 iteration steps, each including one policy update phase and one roll-out phase of 250 timesteps leading to \num{25e3} environment steps per iteration.
To keep the influence of the rewards to a small time horizon, a $\gamma$ factor of 0.6 is selected.
The agent network is updated for ten epochs in the policy update phase.
To account for different image sizes, we use normalized image coordinates as input to the \textit{Variable Encoder}.
\textbf{We will release our code upon acceptance} to provide complete information about our implementation and to facilitate future work.

\section{Deep Agent Ablation}
\label{sec:rationale}
To evaluate the benefits of our  \textit{Variable Encoder} processing the keypoints, we report in Tab.~\ref{tab:euroc_ablation_sup} and Tab.~\ref{tab:tum_ablations} the performance of an agent network without the Variable Encoder, i.e., RL SVO (w/o keypoints).
As can be observed, the agent without access to the keypoint information is not capable of successfully finishing all the sequences of the EuRoC dataset.
%
Furthermore, on TUM-RGBD, the agent without keypoint information only achieves a performance comparable to the heuristic baseline, cf. SVO in the first row of Tab~\ref{tab:tum_ablations}.
This shows that the keypoint information is important for the VO task.

\begin{figure}[ht!]
\centering
\includegraphics[width=0.6\textwidth]{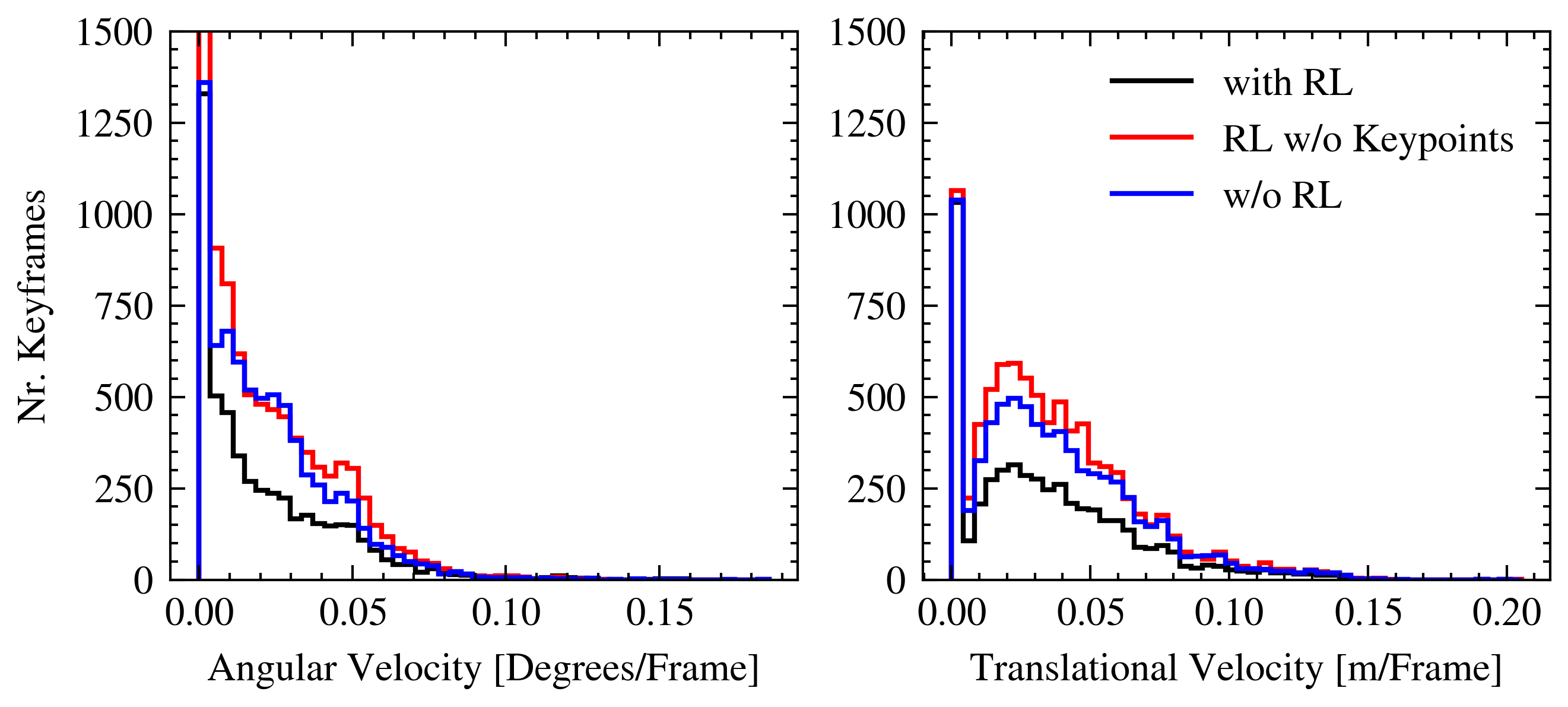}
\caption{
\textbf{Keyframe Selection Ablation.}
The number of selected keyframes at different translational and angular velocities for the EuRoC dataset.
}
\label{fig:sup_keyframe}
\end{figure}

We illustrate in Fig.~\ref{fig:sup_keyframe} the distribution of the selected keyframes over the angular and translational velocities for three variants: our default RL SVO agent, the heuristic-based SVO, and the RL agent without keypoint information.
This figure shows that the RL w/o keypoints features a similar keyframe distribution to the heuristic-based SVO.
The similarity can be explained by the access to the same information for the keyframe selection.
In contrast to the heuristic rules, since our agent is a neural network, we can process complex information without being limited by heuristic-based rules.
Thus, we propose using a Variable Encoder to process a variable number of keypoints.
This architecture leads to a more robust and accurate pose estimation and, as a consequence of the more efficient keyframe selection strategy,  reduces the runtime.
Furthermore, we report in Tab.~\ref{tab:euroc_ablation_sup} and Tab.~\ref{tab:tum_ablations} the performance of the agent trained without privileged critic referred to as RL SVO (w/o privileged c.).
Both agents trained with and without privileged critic can track the same number of sequences on EuRoC.
However, on the TUM-RGBD dataset, the agent trained without privileged critic only finishes 6/9 versus 8/9 achieved by our proposed agent.
This can be explained by the fact that the training using the privileged critic is more stable in terms of the found action distribution, which also leads to a more robust agent.

\begin{table}[ht!]

\newcommand{\cw}{1.1cm}
\centering
\scalebox{0.65}{
\begin{tabular}{m{3.9cm}C{\cw}C{\cw}C{\cw}C{\cw}C{\cw}C{\cw}C{\cw}C{\cw}C{\cw}C{\cw}C{\cw}>{\centering\arraybackslash}m{\cw}}

 Method  & MH01 & MH02 & MH03 & MH04 & MH05 & V101 & V102 & V103 & V201 & V202 & V203 & Avg  \\
  
\hline
SVO           & 0.183 & 0.490 & 0.381 & 3.236 & 1.293 & 0.320 & 0.333 &  -    & 0.084 & 0.553 & 1.431 & - \\
RL SVO                & 0.236 & 0.124 & \textbf{0.241} & 2.149 & 1.982 & 0.225 & 0.398 & \textbf{0.394} & \textbf{0.069} & 0.334 & 1.506 & \textbf{0.696} \\
\hline
RL SVO (w/o keypoints)       & 0.255 & 0.160 & 0.419 & \textbf{1.979} & \textbf{0.588} & \textbf{0.173} & 0.644 & 0.849 & 0.182 & 1.394 & - & -\\
RL SVO (w/o privileged c.)   & \textbf{0.130} & \textbf{0.099} & 0.834 & 3.283 & 0.958 & 0.140 & 0.286  & 0.459 & 0.116 & 0.618 & \textbf{1.373} & 0.754 \\
\hline
RL SVO (high penalty) & 0.156 & 0.148 & -     & -     & 4.349 & 0.245 & 1.244 & 1.251 & 0.109 & 1.715 & 1.473 & - \\
RL SVO (no penalty)   & 0.397 & 0.183 & 0.296 & 2.742 & 1.244 & 0.377 & 0.446 & -     & 0.331 & 0.304 & 1.468 & - \\
\hline
RL SVO (w/o keyframe) & 0.179 & 0.351 & 0.302 & 3.072 & 1.021 & 0.325 & -     & -      & 0.179 & \textbf{0.258} & 1.464 & - \\
RL SVO (w/o gridsize) & 0.152 & 0.183 & 0.293 & 2.154 & 1.537 & 0.198 & \textbf{0.269} & 0.808 & 0.108 & 1.683  & 1.405 & 0.799 \\
\end{tabular}}
\caption{
\textbf{EuRoC Ablations.}
Ablation studies on inputs, reward design, and action of the RL agent.
}
\label{tab:euroc_ablation_sup}
\end{table}
\begin{table}[ht!]

\newcommand{\cw}{1.1cm}
\centering
\scalebox{0.75}{
\begin{tabular}{m{3.9cm}C{\cw}C{\cw}C{\cw}C{\cw}C{\cw}C{\cw}C{\cw}C{\cw}C{\cw}>{\centering\arraybackslash}m{\cw}}

 Method  & 360 & desk & desk2 & floor & plant & room & rpy & teddy & xyz & Avg  \\
\hline
SVO            & \textbf{0.186} & 0.681          & 0.898          & - & 0.320          & 0.805          & \textbf{0.053} & 0.769          & \textbf{0.057}       & 0.471 \\
RL SVO                     & 0.189          & \textbf{0.556} & \textbf{0.755} & - & 0.260          & 0.804          & 0.056          & 0.697          & 0.062 & \textbf{0.422}  \\
\hline
RL SVO (w/o keypoints)     & 0.191          & 0.709          & 0.759          & - & 0.529          & 0.771          & 0.058          & \textbf{0.673}          & 0.095 & 0.473  \\
RL SVO (w/o privileged c.) & -              & 0.635          & 0.821          & - & 0.299          & -              & 0.054          & 0.716          & 0.062 & -  \\
\hline
RL SVO (high penalty)      & -              & 0.575          & 0.906          & - & 0.471          & \textbf{0.723} & 0.059          & 0.738          & 0.087 & -  \\
RL SVO (no penalty)        & -              & 0.611          & 0.900          & - & 0.412          & 0.794          & 0.054          & 0.743          & 0.154 & -  \\
\hline
RL SVO (w/o keyframe)      & -              & 0.652          & 0.891          & - & \textbf{0.243} & 0.841          & -             & 0.711           & 0.079 & - \\
RL SVO (w/o gridsize)      & -              & 0.612          & \textbf{0.755} & - & 0.255          & 0.822          & -             & 0.824           & 0.059 & - \\
\end{tabular}}
\caption{
\textbf{TUM-RGBD Ablations.}
Ablation studies on inputs, reward design, and action of the RL agent.
}
\label{tab:tum_ablations}
\end{table}
\begin{table}[ht!]
\centering
\scalebox{0.75}{
\begin{tabular}{m{3.9cm}>{\centering\arraybackslash}m{2.8cm}}
Method  & Time [ms]\\

\hline
SVO                  &  9.06 \\
RL SVO               &  7.40 \\
RL SVO w/o penalty   & 13.44 \\
RL SVO high penalty  &  6.42 \\
\hline
Network inference    &  1.75 \\
\end{tabular}}
\caption{
\textbf{Inference Runtime.}
The runtime of SVO using different reward weightings.
}
\label{tab:sup_runtime}
\end{table}
\begin{figure}[ht!]
\centering
\includegraphics[width=0.55\textwidth]{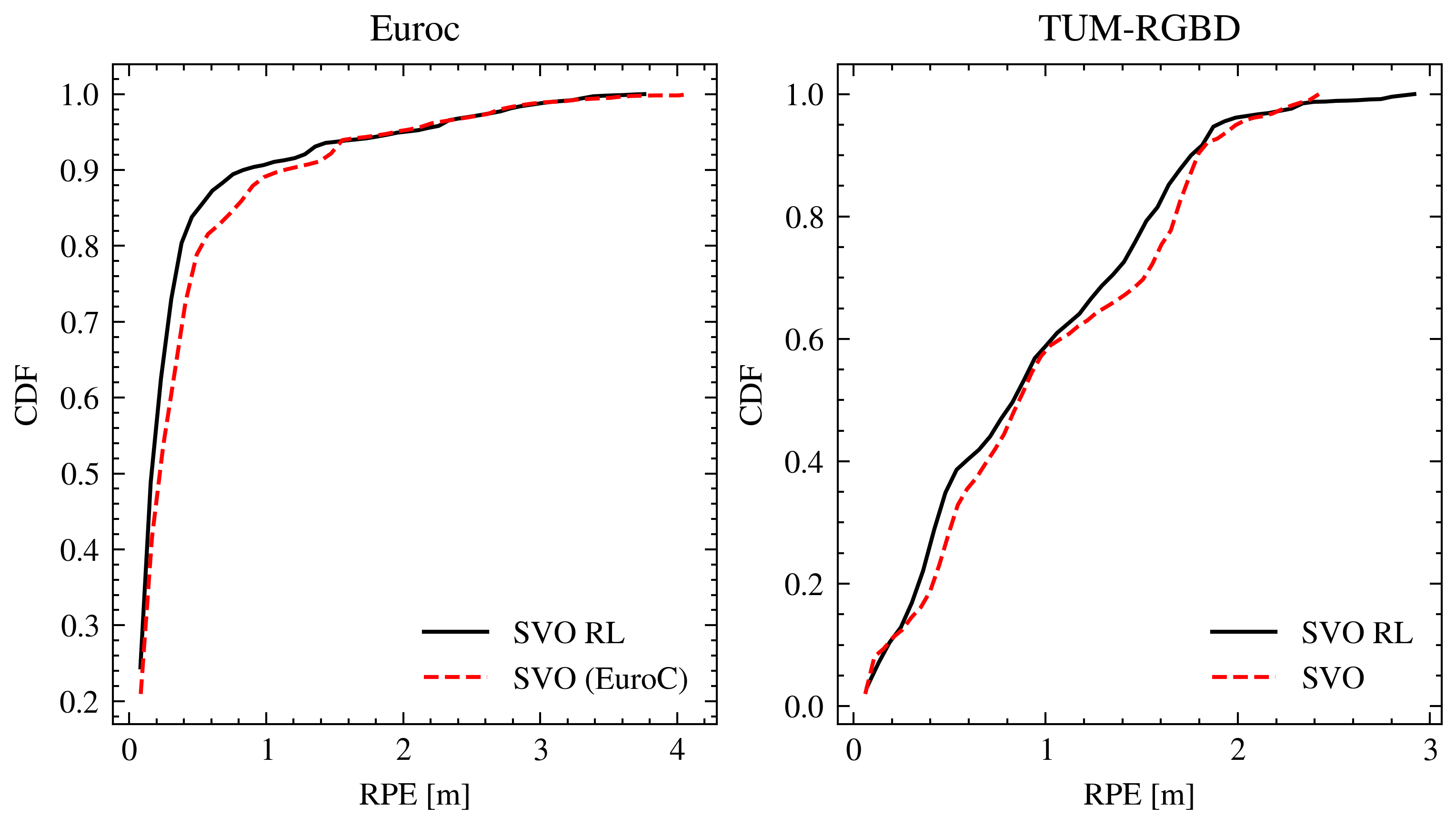}
\caption{
\textbf{Cumulative density function of the relative position error (RPE)}.
The CDF is computed on a sliding window of size 5 m, on the entire test set. 
The steeper the plots are, the better the performance. 
The curve corresponding to the VO with RL is above the curve of the VO without RL.
}
\label{fig:sup_cdf}
\end{figure} 
\begin{table}[!ht]
\newcommand{\cw}{1.5cm}
\centering
\scalebox{0.7}{
\begin{tabular}{m{2.1cm}C{\cw}C{\cw}C{\cw}C{\cw}C{\cw}C{\cw}>{\centering\arraybackslash}m{\cw}}
Method  & 05 & 06 & 07 & 08 & 09 & 10 & Avg \\
\hline
SVO           & 11.91 & 13.31 & \textbf{9.24} & - & \textbf{12.30} & 12.50 & 11.85 \\
RL SVO        & \textbf{11.59} & \textbf{11.12} & 10.42 & - & 12.58 & \textbf{12.02} & \textbf{11.54} \\
\hline
DSO           & 2.42 & - & 2.70 & 3.23 & - & 1.79 & 2.53 \\
RL DSO        & \textbf{2.24} & - & \textbf{2.52} & \textbf{2.92} & - & \textbf{1.74} & \textbf{2.35} \\
\end{tabular}}
\caption{
\textbf{KITTI.}
We report the ratio between ATE and traveled distance in [\%].
Our RL agents achieve lower average error than the heuristic-based VO methods.
}
\label{tab:kitti_results}
\end{table}

\section{Reward Ablation}
\label{sec:reward_ablation}
In Tab.~\ref{tab:euroc_ablation_sup} and Tab.~\ref{tab:tum_ablations}, we also report the accuracy for two variants of our RL agent trained with different weighting on the keyframe action $a_{\text{keyframe}}$, for which we report the runtimes in Tab.~\ref{tab:sup_runtime}.
The first agent RL SVO (no penalty), was trained without any penalization of the keyframe action, which led to an agent selecting almost each frame as a keyframe. 
That also explains the lower accuracy as well as robustness (10/11 successful sequences on EuRoC and 7/9 on TUM-RGBD) compared to the proposed agent with keyframe penalty since a spatially diverse set of keyframes improves the triangulation accuracy and, as a consequence, also the keypoint tracking.
If the penalty on the keyframe action is increased by setting $\lambda_2$ to \num{7.5e-3} (RL SVO high penalty), the agent selects roughly two times fewer keyframes than our default agent. 
Since fewer keyframes are selected, the agent is more prone to lose track due to a small number of keypoints, as happened for sequences MH03 and MH4, see RL SVO (high penalty) in Tab.~\ref{tab:euroc_ablation_sup}.
In general, for both weighting experiments, the training was more unstable than our agent trained with the default value of 0.5 for $\lambda_2$.
As a consequence, the best-performing agent for both trainings with different weighting was already obtained after 150 iterations.

\section{Relative Errors}
In addition to the ATE reported in the main manuscript, we also visualize the relative position error by reporting the cumulative density function (CDF) on the entire dataset, EuRoC and TUM-RGBD, for SVO.
The CDF plot in Fig.~\ref{fig:sup_cdf} shows the probability of the relative position error being below a given value on the x-axis.
The relative position error is computed based on a sliding window of length 5m.
As can be observed, our RL SVO achieves constantly a better performance over the distribution of error distances.

\section{KITTI}
Additionally to EuRoC and TUM-RGBD, we evaluate on six commonly used sequences from the KITTI dataset, recorded by a car and accompanied by poses obtained with an accurate GPS-based localization system.
Considering the long length of the trajectories, we report the error for the KITTI dataset with the predominantly used ratio between ATE and traveled distance in [\%].
Similar to EuRoC and TUM-RGBD, our RL agents also outperform the two heuristic-based VO pipelines specifically tuned for the KITTI dataset, as indicated by the lower mean error in Tab.~\ref{tab:kitti_results}. 
%


%
%
\bibliographystyle{splncs04}
\bibliography{main}
\end{document}